\documentclass[letterpaper, 10 pt, conference]{ieeeconf}  

\usepackage{hyperref}
\IEEEoverridecommandlockouts                              

\usepackage{graphics} 
\usepackage{epsfig} 
\usepackage{mathptmx} 
\usepackage{times} 
\usepackage{amsmath} 
\usepackage{amssymb}  
\usepackage{verbatim} 
\usepackage{textcomp}
\usepackage{url}
\usepackage{booktabs,chemformula}

\usepackage{pifont}
\definecolor{applegreen}{rgb}{0.55, 0.71, 0.0}
\newcommand{\cmark}{\textcolor{applegreen}{\ding{51}}}%
\newcommand{\xmark}{\textcolor{red}{\ding{55}}}%

\newcommand{\hand}{BaRiFlex}

\usepackage{caption}
\usepackage{subcaption}
\usepackage{siunitx}
\usepackage{tikz}

\usepackage[backend=biber,
            hyperref=true,
            url=false,
            isbn=false,
            doi=false,
            backref=false,
            style=ieee,
            citestyle=numeric-comp,
            sorting=nyt,
            block=none]{biblatex}
\usepackage[font=small]{caption}

\newbox{\bigpicturebox}

\title{\LARGE \bf
BaRiFlex: A Robotic Gripper with Versatility and Collision Robustness for Robot Learning
}

\author{Gu-Cheol Jeong$^{1}$, Arpit Bahety$^{2}$, Gabriel Pedraza$^{1}$, Ashish D. Deshpande$^{1}$ and Roberto Mart{\'i}n-Mart{\'i}n$^{2}$
\\
\thanks{*This work was not supported by any organization}
\thanks{$^{1}$Gu-Cheol Jeong, Gabriel Pedraza, and Ashish D. Deshpande are with the Department of Mechanical Engineering, The University of Texas at Austin, USA.
        {\tt\small goochul@utexas.edu}}%
\thanks{$^{2}$Arpit Bahety and Roberto Martin-Martin with the Department of Computer Science, The University of Texas at Austin, USA.}%
}

\begin{document}

\maketitle
\thispagestyle{empty}
\pagestyle{empty}


\begin{abstract}
We present a new approach to robot hand design specifically suited for successfully implementing robot learning methods to accomplish tasks in daily human environments. 
We introduce \hand{}, an innovative gripper design that alleviates the issues caused by unexpected contact and collisions during robot learning, offering robustness, grasping versatility, task versatility, and simplicity to the learning processes. This achievement is enabled by the incorporation of low-inertia actuators, providing high \underline{Ba}ck-drivability, and the strategic combination of \underline{Ri}gid and \underline{Flex}ible materials which enhances versatility and the gripper’s resilience against unpredicted collisions. Furthermore, the integration of flexible Fin-Ray linkages and rigid linkages allows the gripper to execute compliant grasping and precise pinching. We conducted rigorous performance tests to characterize the novel gripper's compliance, durability, grasping and task versatility, and precision. We also integrated the \hand{} with a 7 Degree of Freedom (DoF) Franka Emika’s Panda robotic arm to evaluate its capacity to support a trial-and-error (reinforcement learning) training procedure. The results of our experimental study are then compared to those obtained using the original rigid Franka Hand and a reference Fin-Ray soft gripper, demonstrating the superior capabilities and advantages of our developed gripper system. More information and videos at\\{\url{https://robin-lab.cs.utexas.edu/bariflex}}

\end{abstract}


\section{Introduction}
\label{s:intro}

\begin{figure}
    \centering
    \includegraphics[width=1\columnwidth]{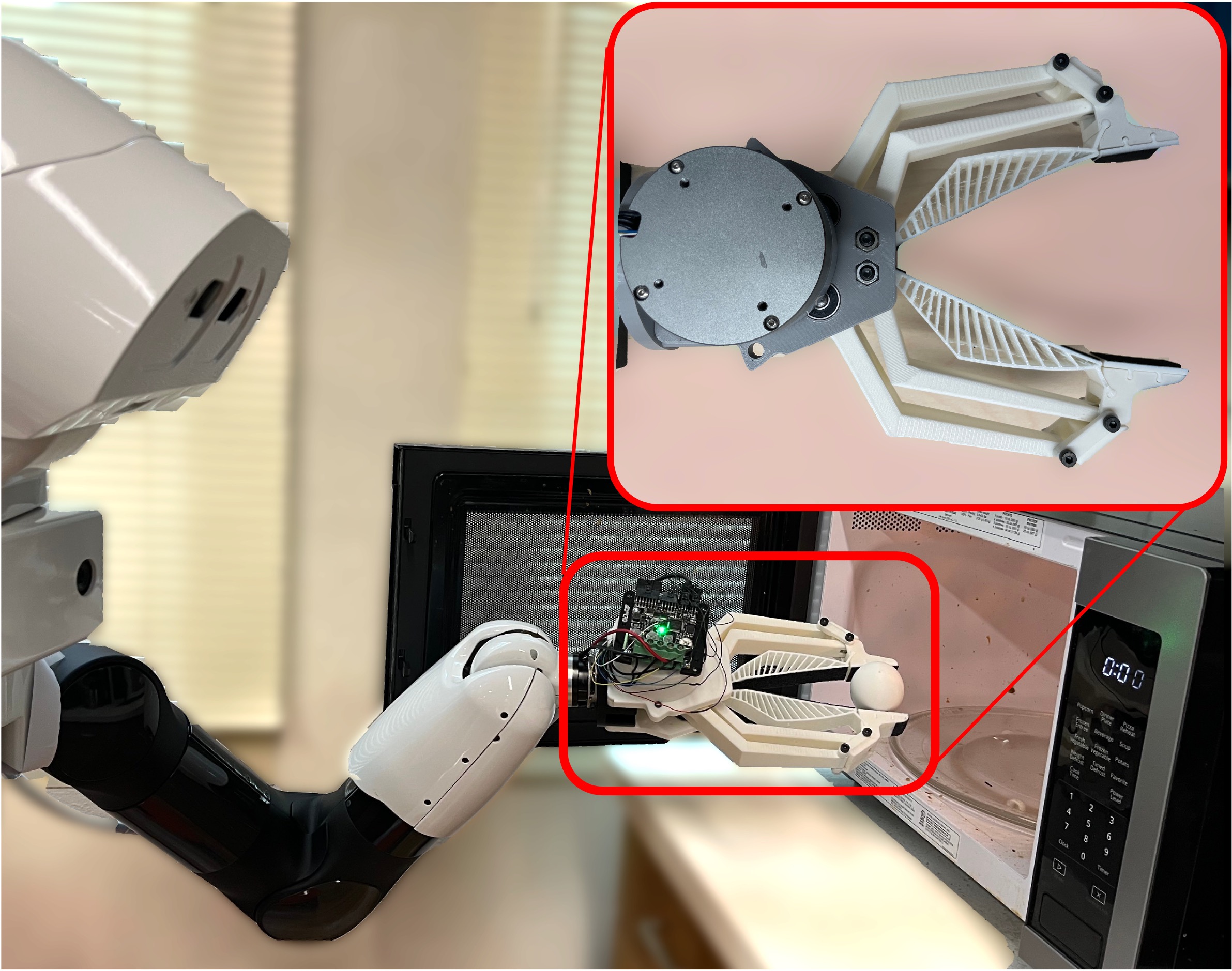}
    \caption{\hand~mounted on a robot arm and holding a delicate object (an egg). The novel combination of design features in \hand{} -- high \underline{Ba}ck-drivable actuator and hybrid \underline{Ri}gid-\underline{Flex}ible structure -- provides robustness through compliance against contact interactions, adaptability and versatility for grasping, precision for pinch grasps, and support to manipulate heavy objects, enabling a variety of manipulation tasks in unstructured environments. All with a low cost and simple actuation, characteristics that facilitate robot learning.}
    \label{fig:GripperArm}
\end{figure}


Robots manipulating in unstructured environments must deal with the environment's inherent uncertainty and a large variability of objects and tasks.
For example, a service robot performing an everyday activity such as \textit{preparing a cup of instant coffee} will face a different kitchen in every home and have to manipulate multiple objects in different ways: picking and placing them, opening/closing their articulation, pressing their buttons\ldots
This required generalization and versatility are challenges that have impeded the development of household robots for decades.

In recent years, the use of robot learning techniques such as reinforcement learning~\cite{sutton2018reinforcement}, or imitation learning~\cite{schaal1999imitation,mandlekar2021matters} provided more capable and versatile solutions for robots to manipulate unstructured environments~\cite{kartmann2021semantic,wu2023tidybot,brohan2023can,brohan2022rt}.
However, robot learning comes with costs including extensive trial and error, initial inaccuracies during training, and frequent collisions with the environment~\cite{peters2016robot,kroemer2021review}.
These costs can result in irreversible damage to the robot, especially when it's equipped with traditional rigid hands to interact, hands that are often also difficult to exploit for learning due to their high complexity and large number of active degrees of freedom~\cite{bekey1998autonomous,brooks1993real,zakka2023robopianist,andrychowicz2020learning}.

To deal with these challenges in robot learning, the community has resorted to embodiments that can handle unexpected contact through compliance and simplify actuation, for example through underactuated~\cite{maYaleOpenHandProject2017, yoon_fully_2022, Odhner2014, jeong_design_2019, Santina2018, shadowhand} or flexible hands~\cite{noauthor_adaptive_nodate, puhlmannRBOHandPlatform2022, Santina2018}.
However, these previous hand designs led to narrower task versatility: grasping some necessary objects and performing some tasks became unfeasible.


In this work, we present a novel hand, \hand{} (Fig.~\ref{fig:GripperArm}), designed to support the requirements of robot learning for manipulation in unstructured environments, namely the robustness to resist frequent and unexpected contacts, and the versatility to grasp and execute the different tasks involved in activities such as household chores.
These essential characteristics are achieved through a combination of 1) high back-drivability facilitated by a low-inertia actuator, and 2) high adaptation through a hybrid design structure that synergistically combines both rigid and flexible components. 
Our approach not only absorbs collisions protecting the robot arm from high contact forces, but also provides the necessary flexibility to enable delicate and fine-grained manipulation of small objects while maintaining the ability to apply large forces to grasp and/or interact with heavy objects and to move extremely fast to react to dynamic events. All with a low budget (under 500 USD), and simple 3D printed elements (under 38 hours) and assembly (under 1 hour).

We perform an extensive experimental evaluation to characterize physically and functionally our new \hand{} hand design. Our experiments indicate that the high compliance, durability, adaptation, and precision of \hand{} facilitates grasping a large variety of everyday objects, and supports performing tasks extending beyond simple object grasping such as pressing buttons and opening door handles. Moreover, \hand{} also enables learning through continuous trial-and-error in a real-world reinforcement learning loop, all without damaging the mechanism despite several collisions. 


The foremost takeaway from this paper is the crucial role played by the novel \hand{} gripper in overcoming the hardware constraints that have previously limited the applicability of robot learning. 
The design of the gripper is motivated by our analysis of the requirements for hands to enable robot learning in human daily environments that we also contribute here (Sec.~\ref{sec:requirements}). 
The result is the fusion of high robustness and versatility in a single gripper, that empowers robotic agents to successfully perform an array of tasks that conventional grippers couldn't achieve.



\section{Requirements in a Robot Hand for Learning}
\label{sec:requirements}

\newcommand{\high}{\cmark\cmark}
\newcommand{\medium}{\cmark}
\newcommand{\low}{\xmark}

\begin{table*}[h]
  \scriptsize
  \centering
  \caption{Comparison of grippers and hands in terms of Structure, Features, and Attributes}
  \begin{tabular}{@{}c||ccc|cccc|cccc@{}}

    \toprule
     &  \multicolumn{3}{c}{\textbf{Structure}} & \multicolumn{4}{|c|}{\textbf{Features}} & \multicolumn{4}{c}{\textbf{Attributes}}\\
    Gripper  & Actuation & Transmission & Material & Payload & Speed & \begin{tabular}{@{}c@{}}Back- \\ Drivabiltiy\end{tabular}  & Compliance & Robustness & GV & TV & Simplicity \\
    \midrule
    BLT Gripper~\cite{kimBLTGripperAdaptive2020} & Geared & \begin{tabular}{@{}c@{}}Linkage \& \\ Lead Screw\end{tabular} & Rigid\&Soft & 10N & \medium{} & \medium{} & 
 \medium{} & \medium{}& \high{} & \high{} & \medium{}\\
    Robotiq~\cite{robotiq_2F-140}& Geared & Linkage & Rigid & 10-125N & \medium{} & \medium{} & \medium{} & \medium{} & \medium{} & \high{}& \high{}\\
    Magripper~\cite{Tanaka2020} & Direct-Drive & Linkage & Rigid & NM & \high{} & \high{} & \high{} & \high{}& \low{} & \low{} & \high{}\\
    Open Hand~\cite{maYaleOpenHandProject2017}& Geared & Tendon & Rigid  & 9.6N & \medium{} & \medium{} & \medium{} & \medium{}& \medium{} & \medium{} & \high{}\\
    DD Hand~\cite{Bhatia2019} & Direct-Drive  & Linkage & Rigid  & 6N & \high{} & \high{} & \high{} & \high{}& \low{} & \medium{} & \medium{}\\
    Allegro Hand~\cite{bae_development_2012}& Geared & Linkage & Rigid  & 50N & \high{} & \low{} & \low{} & \low{} & \high{} & \high{} & \low{} \\    
    RBO Hand3~\cite{puhlmannRBOHandPlatform2022} & Pneumatic & Tube & Soft  & 8.3N & \high{} & \high{} & \high{} & \high{} & \high{} & \high{} & \low{}\\
    \begin{tabular}{@{}c@{}}Fin-Ray\textregistered ~\cite{crooks_fin_2016} \\ \& Franka Hand\end{tabular} & Geared & NM & Soft  & NM & \medium{} & \medium{} & \high{} & \high{} & \medium{} & \medium{} & \high{}\\
    Franka Hand~\cite{frankahand} & Geared & NM & Rigid  & 70-140N & \medium{} & \low{} & \low{} & \low{} & \low{} & \low{} & \high{}\\
    \hand & Direct-Drive & Linkage & Rigid\&Soft  & 11N & \high{} & \high{} & \high{} & \high{} & \high{} & \high{} & \high{}\\
    \bottomrule
  \end{tabular}
  \label{table:Comparison_table}\\
NM=Not Mentioned, GV=Grasping Versatility, TV=Task Versatility, \high{}=high, \medium{}=medium, \low{}=low
\end{table*}

The human hand is a remarkable manipulation apparatus, not just due to its ability to handle both heavy and small objects but also because of its diverse range of functional capabilities and its robustness in adapting and absorbing impact~\cite{gustus2012human,schwarz1955anatomy,napier1956prehensile,jones2006human}. 
In contrast, most robot hands are hard and rigid, possibly inherited from the design of robots in structured factory domains, which lead to brittle and dangerous interactions when facing unexpected contact~\cite{bae_development_2012, kimBLTGripperAdaptive2020, maYaleOpenHandProject2017, robotiq_2F-140, frankahand, sandiahand, Onrobot, WeissRobotics, cerruti_alpha_2016,brusa_mechatronics_2015,franchi_baxter_2015,ceccarelli_design_2017,ciocarlie_velo_2014,melchiorri_development_2013,ulrich_medium-complexity_1988,salisbury_articulated_1982}.
Humans, supported by the versatility and robustness of their hands, learn and demonstrate superior manipulation abilities by leveraging the same contact that robots with rigid hands try to avoid.
A look at a human performing household activities~\cite{damen2018scaling,bullock2015yale,grauman2022ego4d} gives us an intuition of the rich use of physical interaction between the hands and the environment and objects. 
For example, when preparing a cup of instant coffee, a human would use their hands to open heavy objects such as the fridge, pick large objects such as the milk or the coffee jar but also small objects such as the spoon and delicate objects such as the glass, and press buttons in the appliances such as the microwave control and opening mechanism; a large variety of objects and tasks, just in a single activity!
These skills are acquired through trial-and-error, beginning in infancy. Research with infants shows that they extensively utilize contact interactions to learn how to interact with the world~\cite{smith2005development,hoffer1947mouth,adolph2017development,bourgeois2005infant,cangelosi2018babies}.
Prior work in robotics evidenced the benefits of enabling similar rich interactions in robots~\cite{agrawal2016learning,bohg2017interactive,martin2022coupled}.

Based on these observations, we would like to equip our robots with a hand that possesses the specific attributes that support learning and manipulation tasks in unstructured environments. 
The attributes we have identified include 1) an innate adaptation to contact and collisions, 2) proficiency in grasping objects of diverse shapes and sizes, and 3) the potential to support a wide range of tasks beyond simple grasping motions, all with 4) the simplicity in the manufacturing and actuation to empower current learning algorithms.
In the following, we provide a working definition of these attributes. In the next section, we will analyze previous hand designs based on them.


\subsubsection{\textbf{Robustness}}
A hand exhibits robustness through its \textit{capability to resist collision and unexpected interactions with the environment without breaking or losing functionalities}.
Robustness could be achieved with a hard and strong hand, but this would transmit the impact of the collision to the robot arm, potentially causing damage to both the hand and the robot.
An alternative is to create \textit{compliance} in the actuator, through the mechanism, the materials, or both, so that the impact forces are partially absorbed by the hand, protecting the arm.
In this case, to be robust a compliant hand should provide some form of elastic behavior: when the collision disappears, the hand recovers its original capabilities without damage.

\subsubsection{\textbf{Grasping Versatility}}
Hands and grippers with high grasping versatility have the \textit{potential to grasp a wide range of objects} of different sizes, shapes, masses, surface materials, and other physical properties.
This is frequently attained by hands that can perform multiple types of grasps~\cite{feix2015grasp} such as power and pinch grasps, and/or by hands that can conform to the shape of the objects, and succeed even when the assumed object pose is inaccurate.
Grasping versatility is linked to the robot's application domain: being able to grasp all objects involved in the application domain (e.g., household activities) is a prerequisite for its success.

\subsubsection{\textbf{Task Versatility}}
Robots in unstructured environments need to perform multiple manipulation tasks: picking, pushing, opening, turning, catching\ldots
Task versatility is the \textit{capability of a hand to support a wide range of operations beyond grasping}.
This capability is backed by various properties of hands such as actuation speed, size, hardness, or accuracy in its movement.
For example, pressing a button may require the fingertips to be compact and strong enough, catching a falling object requires the robotic hand to have a high actuation speed, opening a heavy door needs a hand that can transmit enough force, and motion accuracy may be necessary to manipulate delicate objects. 
As with grasping versatility, task versatility is scoped by the robot's application domain: a robot hand intended to enable household activities should demonstrate that is capable of performing all tasks related to that domain.

\subsubsection{\textbf{Simplicity}}
The human hand is a highly complex mechanism with 27 semi-coupled degrees of freedom controlled by over 30 muscles. 
In an attempt to achieve human dexterity, some robot hands have tried to emulate this complexity~\cite{deshpande_mechanisms_2013, shadowhand}.
However, the current state of computation and learning is bounded, and some of the most successful solutions to manipulation have been based on simple actuation~\cite{levine2018learning,brohan2023can}. 
While in the future higher levels of grasping and task dexterity may be enabled by complex hands, the current state of robot learning demands simple hand designs.
Simplicity in a robotic hand relates to an \textit{uncomplicated design and clean mechanism}, usually with few active degrees of freedom and straightforward construction. 
Additionally, complexity directly impacts price and the difficulty of repairing and maintaining the hand; a simple design is usually linked to a lower price and quick and easy fixes.

In the next section, we will use the hand attributes defined above to analyze previously proposed hand designs.

\section{Related Work}
\label{s:rw}

The literature on hand designs is vast and rich. In the following, we analyze the most relevant families of gripper designs and their relationship to the attributes discussed above, with Table~\ref{table: Comparison table} presenting a more detailed comparison of some of the most-used robotic hands. 

Most designs for robotic hands follow the same type of construction as robot arms: a series of rigid links connected by actively controlled joints~\cite{bae_development_2012, frankahand, ulrich_medium-complexity_1988,salisbury_articulated_1982}. While providing full controllability, high precision, and potentially high versatility, these hands tend to be expensive, brittle, and highly complex, restricting their success to known structured environments where unexpected contact is reduced and the need for learning is low. To overcome these challenges, researchers have explored semi-rigid underactuated~\cite{kimBLTGripperAdaptive2020, robotiq_2F-140, Tanaka2020} and soft hand designs~\cite{noauthor_adaptive_nodate, puhlmannRBOHandPlatform2022,Santina2018, crooks_fin_2016}. They improve robustness and some of them achieve high grasp versatility thanks to the ability to conform to the objects, but, with a few exceptions, they tend to be complex, with many degrees of freedom and difficult manufacturing processes. Due to their compliant design, many of them lose the capability to perform tasks related to household activities, e.g., the ones involving precision or the application of high forces.
With \hand{}, our goal is to strike a good balance between robustness through compliance, versatility for grasping and for multiple tasks related to the household domain, all with a simple design and low cost.



In Table~\ref{table:Comparison_table} we include the designs more relevant and/or similar to \hand{}. 
The BLT Gripper~\cite{kimBLTGripperAdaptive2020} demonstrates high grasping versatility by combining active compliant belt and rigid linkages. However, its lead-screw transmission with a high-geared ratio makes the gripper brittle and easy to damage due to unexpected collisions.
The Direct Drive Hand~\cite{Bhatia2019} achieved remarkable collision robustness by utilizing two direct drive actuators per finger with a low-friction rigid linkage mechanism, but it lacks the grasping versatility and task versatility, as only pinch motion can be utilized to grasp the object. 
Allegro Hand~\cite{bae_development_2012} stands out as the most universal robotic hand among the fully actuated systems.
Its high DoF improves the task and grasping versatility but increases the learning complexity. Additionally, due to space constraints, high-gear motors were used, which are not back-drivable.
RBO Hand~\cite{puhlmannRBOHandPlatform2022} achieved collision robustness and high grasping versatility by employing soft compliant material at the linkage. 
This soft compliant linkage enhances the contact area through object-adaptive grasping mechanisms.
Despite these advantages, the soft linkage mechanism is less precise than the rigid hand and faces challenges when attempting precise pinching of objects with the fingertips.

In \hand{}, our approach to achieving robustness, task versatility, and grasping versatility involves combining a high back-drivable system with compliant materials (soft materials and torsion springs). 
This combination results in compliance while keeping costs low by using readily available parts, ultimately enhancing the gripper's robustness.
Additionally, we employ a combination of rigidity through a direct drive actuator and flexible mechanism to enable both pinch grasping and adaptable grasping. 
This approach significantly enhances the \hand{}'s versatility when it comes to grasping various objects.
In the next section, we provide a detailed description of the design of \hand{}.

\section{\hand{} Design Concept}
\label{s:design}

\captionsetup[subfigure]{font=small,skip=1pt}
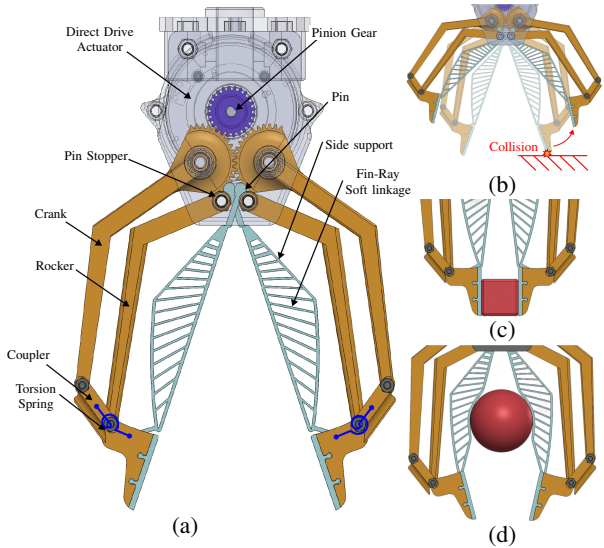
\begin{figure}
    \centering
\sbox{\bigpicturebox}{%
\begin{subfigure}[b]{.29\textwidth}
\begin{tikzpicture}
    \draw (0, 0) node[inner sep=0] {\includegraphics[width=1\textwidth]{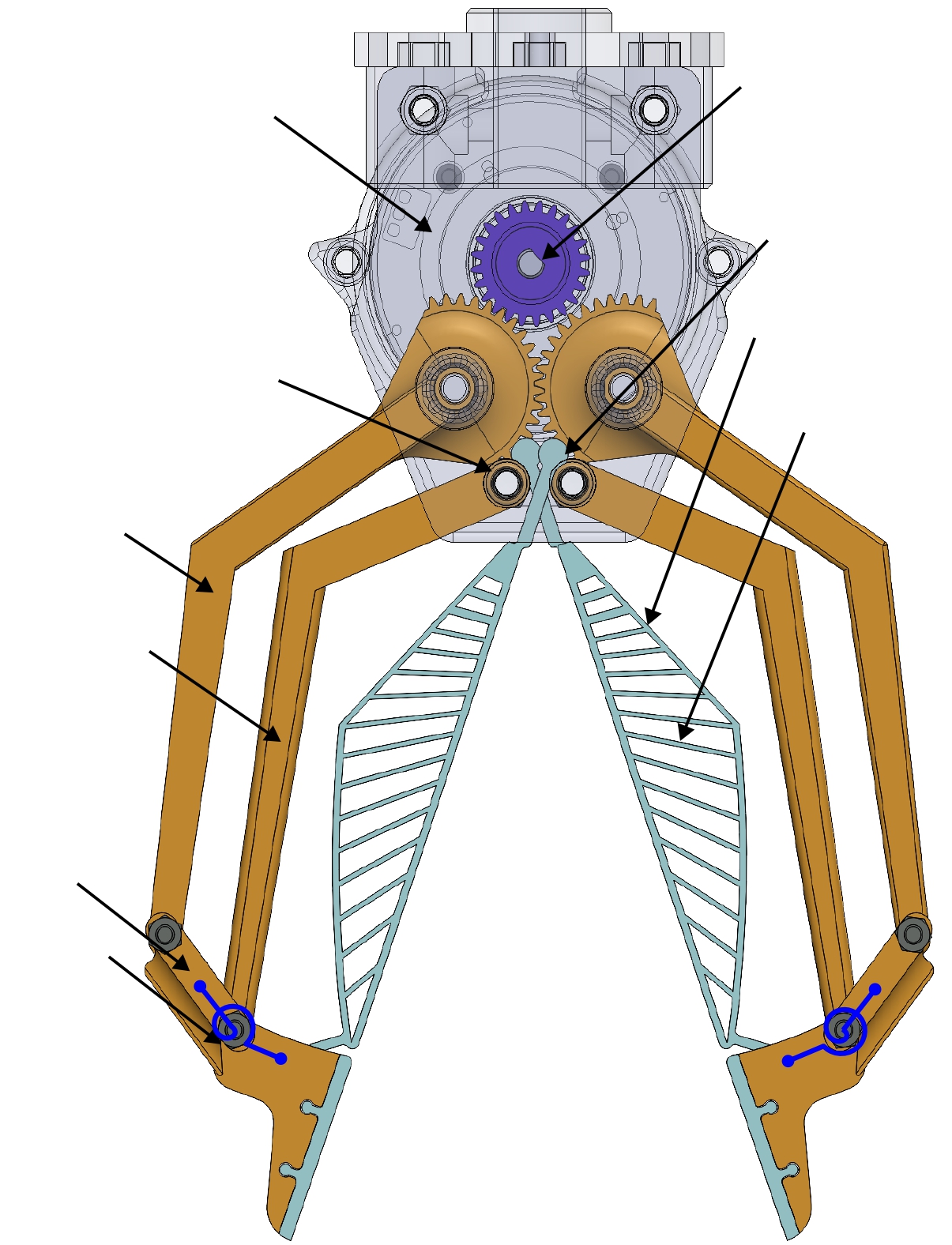}};
    \draw (1.8, 3.0) node [fill=white,inner sep=1pt] {\tiny Pinion Gear};
    \draw (-1.45, 2.9) node {\tiny {\begin{tabular}{c} Direct Drive \\ Actuator\end{tabular}}};
    \draw (2.25, 1.0) node {\tiny {\begin{tabular}{c} Fin-Ray \\ Soft linkage\end{tabular}}};
    \draw (-2.1, 0.60) node {\tiny Crank}; 
    \draw (-2.05, -0.1) node {\tiny Rocker}; 
    \draw (-2.4, -1.3) node {\tiny Coupler}; 
    \draw (-2.3, -1.85) node {\tiny {\begin{tabular}{c} Torsion \\ Spring\end{tabular}}}; 
    \draw (-1.5, 1.35) node {\tiny Pin Stopper};
    \draw (1.75, 2.15) node  {\tiny Pin};
    \draw (2.0, 1.5) node  {\tiny Side support};
\end{tikzpicture}
\caption{}
\end{subfigure}
}%
\usebox{\bigpicturebox}\hfill
\begin{minipage}[b][\ht\bigpicturebox][s]{.18\textwidth}
\begin{subfigure}{.85\textwidth}
\begin{tikzpicture}
    \draw (0, 0) node[inner sep=0]{\includegraphics[width=\textwidth]{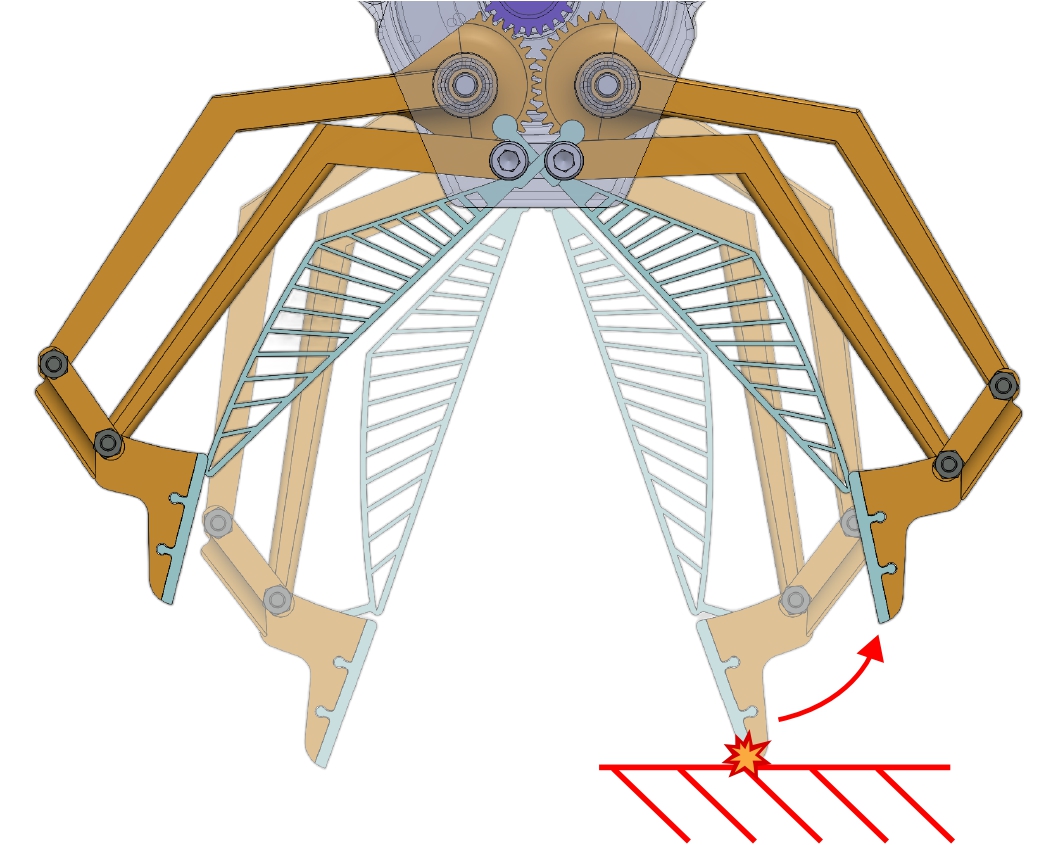}};
    \draw (0.15, -0.78) node[text=red] {\tiny Collision};
\end{tikzpicture}
\caption{}
\end{subfigure}\hfill \\
\begin{subfigure}{.85\textwidth}
\includegraphics[width=\textwidth]{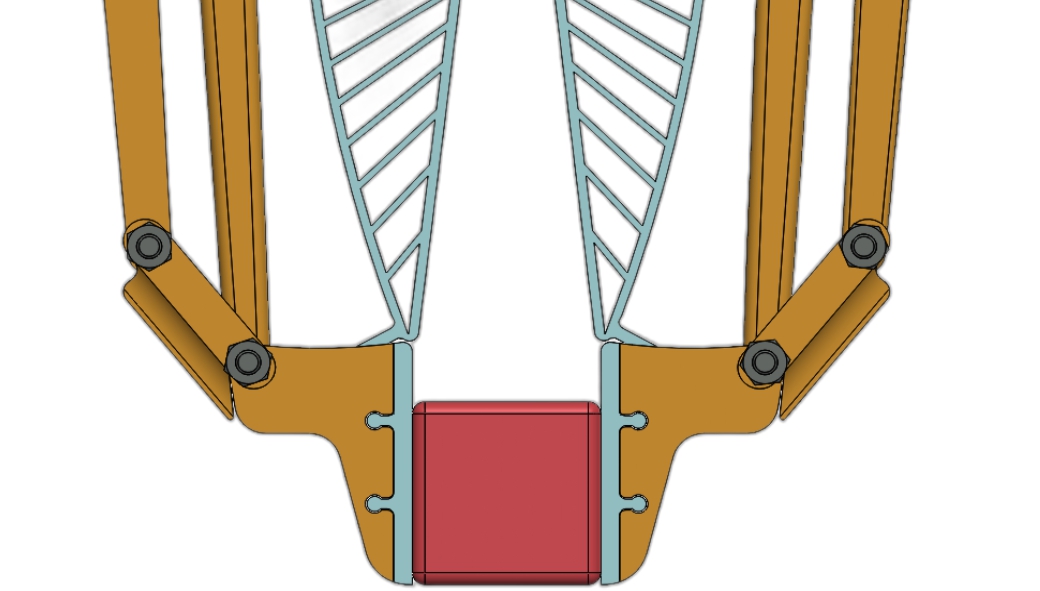}
\caption{}
\end{subfigure}\\
\begin{subfigure}{.85\textwidth}
\includegraphics[width=\textwidth]{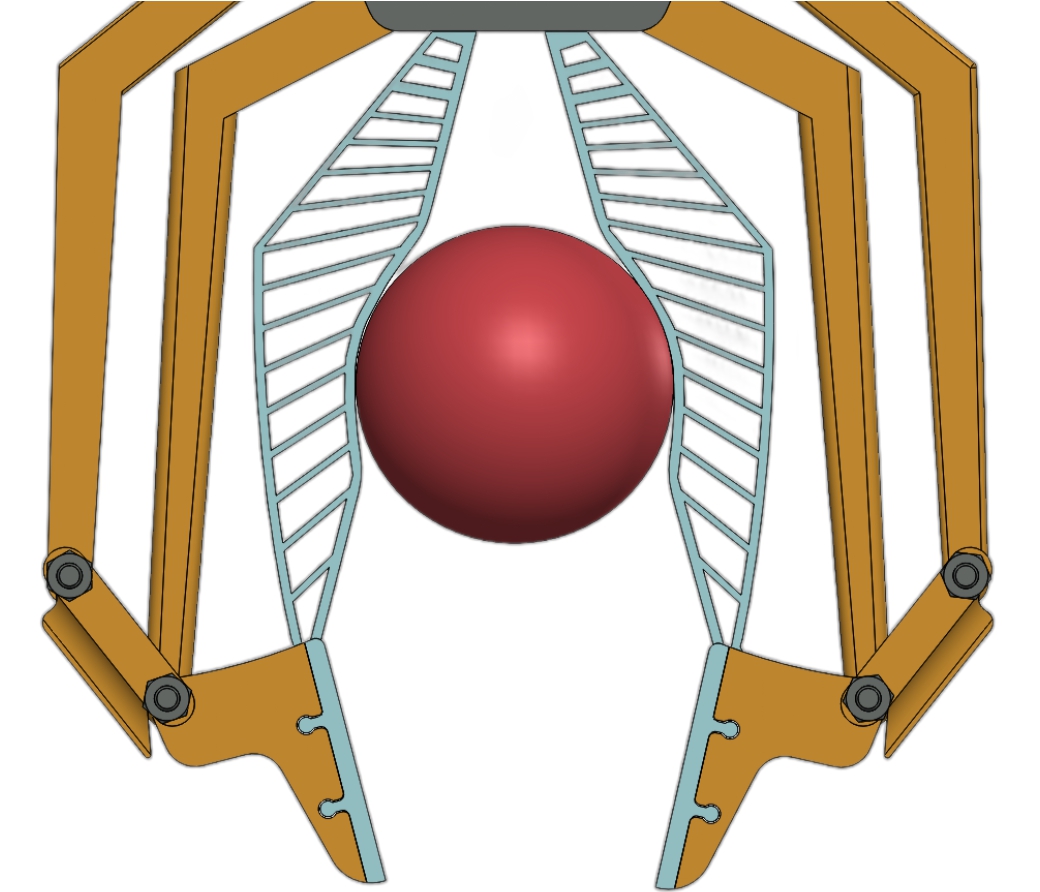}
\caption{}
\end{subfigure}\hfill
\end{minipage}
    \caption{Design structure of the BaRiFlex gripper. (a) The rigid 4-bar linkage mechanism transmits the torques from the direct drive actuator and the collision forces back to it, enabling (b) smooth back-drive motion, and facilitating (c) parallel precise grasping when combined with the underactuated fingertips. The torsional springs at the fingertips further enhance robustness by absorbing collision forces. The inner linkage is constructed with a Fin-Ray structure of soft 3D printed material that yields (d) compliant grasping with adaption to the objects' shape, increasing \hand{}'s grasp versatility. The design is simple, with only one actuated DoF and no gearbox, cost is under 500 USD, and is manufacturable in one day with two 3D printers.
    }
    \label{fig:gripper_mechanism}
\end{figure}

Our design decisions for \hand{} are made to satisfy the requirements for robot learning,  namely, robustness, versatility, and simplicity (Section \ref{sec:requirements}). \hand{} design is simple with only one actuator and weighs under \SI{750}{\gram}. Robustness and versatility are accomplished by a rigid-flexible hybrid mechanism and a high back-drivability mechanism.


\subsection{Rigid-Flexible Hybrid Mechanism}
To enable \hand{} to grasp diverse objects and perform a wide range of tasks while also maintaining robustness, we designed a combination of rigid and flexible mechanisms.

The rigid portion comprises two 4-bar mechanisms (one on each side) with low-friction bearings at the joints, enabling the gripper to achieve efficient power transmission and precise pinch grasping by orienting the fingertips to face each other during the motion. The actuator's pinion is connected to one side of the outer linkage via the main gear, which subsequently rotates to engage the opposite-side finger linkage (Fig.~\ref{fig:gripper_mechanism}a). 
The lengths of all linkages have been carefully selected in order to grasp many daily objects such as a water bottle, mustard container, and cereal box. As a result, we are able to pinch objects with a width of \SI{200}{\milli\meter} at the fingertip and grasp objects with a depth of \SI{60}{\milli\meter} using the inner flexible linkage as presented in Table~\ref{table: Mechanical Specification}. 
The 4-bar mechanism also ensures that the fingertip link has no more than \SI{10} degrees of movement while the gripper is in operation. The fingertip joint possesses a torsion spring that couples the fingertip to the coupler, allowing for compliant flexion and disturbance rejection. 
A mechanical stopper on the coupler not only prevents excessive fingertip extension but also transmits disturbance forces to the actuator.

As depicted in Fig.~\ref{fig:gripper_mechanism}d, the flexible Fin-Ray effect mechanism is employed to achieve compliant grasping motion. 
Unlike prior Fin-Ray grippers, which increase the thickness of the side support for added strength and rigidity~\cite{yao_research_2023} (which reduces flexibility), our gripper incorporates a thin side support structure to facilitate retraction.
We selected thermoplastic polyurethane with a Shore-A hardness of 95 (TPU-95A) due to its flexibility and its ability to effectively transfer support forces from the rocker linkage to the object. 
The TPU Fin-Ray flexible linkage is equipped with a pin at one end that moves freely up and down to prevent any overconstraining by the 4-bar linkage.
Additionally, the inner joint of the rocker linkage serves as a stopper and guide for this TPU pin, ensuring that the pins remain within their intended range.
These flexible Fin-Ray linkages facilitate adaptive grasping with enhanced contact region between the gripper and unknown objects, leading to the application of uniformly distributed force with only one actuator, thereby contributing to grasping versatility.

\begin{table}[t!]
\caption{Mechanical characterization of \hand{}}
\label{table_example}
\begin{center}
\begin{tabular}{|c||c|c|}
\hline
Specification & Value & Unit \\
\hline
Weight & 750 & $g$ \\
Size & 125$\times$255$\times$83 (W$\times$H$\times$T)& $mm$ \\
Max Stroke & 200 & $mm$ \\
\begin{tabular}{@{}c@{}}Continuous Force \\ (Fingertip)\end{tabular} & 11 & $N$ \\
Range of Motion & 86.5 & $Deg$ \\
Speed & 1440 & $Deg/s$ \\
Closing time & 0.18 & $Seconds$ \\
Rated Torque & 0.6 & $Nm$ \\
Gear-Pinion Ratio & 1.54 (37T-24T)& $-$ \\
\hline
\end{tabular}
\end{center}
    \label{table: Mechanical Specification}
\end{table}

\subsection{High Back-drivability Mechanism}
The utilization of a low-friction rigid linkage mechanism ensures that unexpected collision forces can be transferred to the low-inertia actuator. 
This actuator, characterized by its high back-drivability, plays a crucial role in absorbing and mitigating the impact forces generated during collisions.
The T-Motor GL60 brushless DC (BLDC) gimbal motor was chosen due to its ability to provide precise and stable control, along with its compact size and lightweight design. 
To accurately sense the motor's position, an AMS-AS5048A 14-bit absolute encoder was mounted on the motor.
The collision force transfers to the actuator through the rigid linkage mechanism, and the low inertia actuator retracts based on the transferred torque direction.
High speed is an additional feature achieved by the direct drive actuator, allowing for an open-to-close motion in just \SI{0.18} seconds.

\captionsetup[subfigure]{font=small,skip=1pt}
\begin{figure}
\vspace{-0.5cm}
\centering
\begin{subfigure}[b]{.125\textwidth}
\begin{tikzpicture}
    \draw (0, 0) node[inner sep=0] {\includegraphics[width=\textwidth]{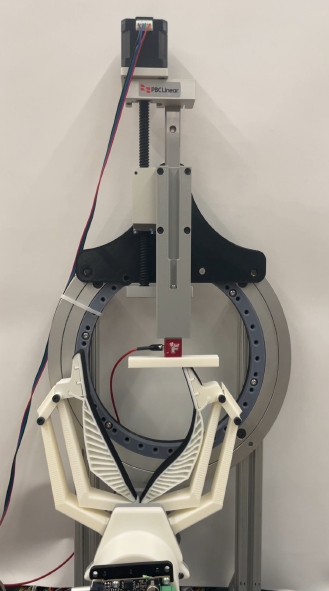}};
    \draw (0.5, 1.5) node [fill=white,inner sep=1pt] {\tiny Stepper Motor};
    \draw (0.7, -0.5) node [fill=white,inner sep=1pt] {\tiny Load cell};
\end{tikzpicture}
\end{subfigure}
\hfill
\begin{subfigure}[b]{.34\textwidth}
        \centering
    \includegraphics[trim={0.5cm 0 1.3cm 0},clip,width=1\columnwidth]{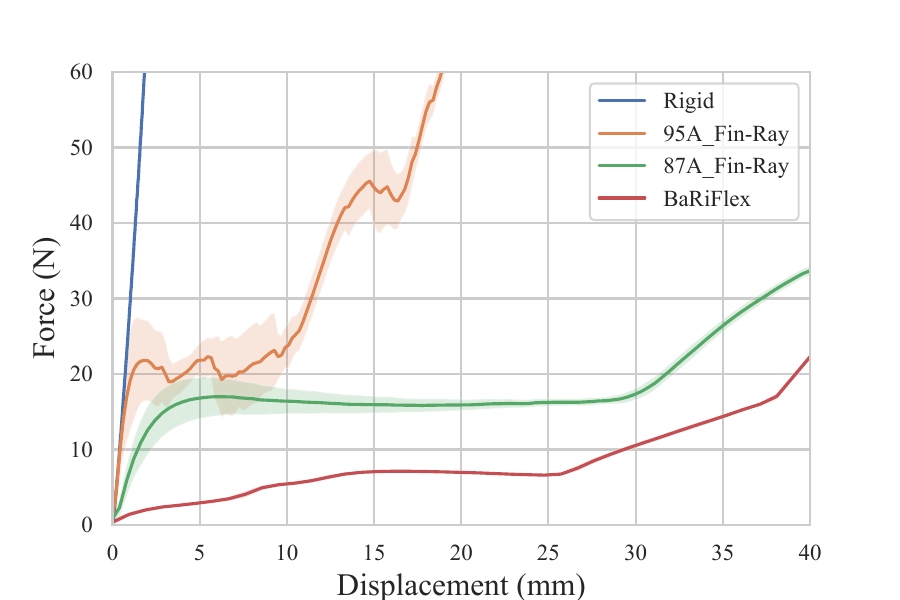}
\end{subfigure}
    \caption{Experimental evaluation for robustness/compliance. An accurate linear stepper motor presses on the tested hand (left). The reactive forces corresponding to different collision distances are recorded (right). \hand{} exhibits the highest compliance, being able to absorb more impact forces, which facilitates interactions and learning in unstructured environments.
    }

    \label{fig:compliance}
\end{figure}


\section{Experimental Evaluation}
\label{s:exps}


Our experiments aim to assess how our gripper design decisions impact its performance within a robot learning context. In particular, we designed the experiments to answer the following three questions: 1) Is the gripper robust enough to tolerate multiple collisions? 2) How versatile is the gripper to grasp different shapes, sizes, and variations in poses of objects? 3) How versatile is the gripper in terms of the tasks that it can enable to perform? Finally, we perform an integrated experiment where we learn with real-world reinforcement learning to grasp an object using the BaRiFlex gripper, evaluating its capabilities to enable robot learning.

\subsection{Evaluating Robustness}

In \hand{}, robustness is obtained through compliance, the ability to absorb impact with an elastic deformation, attained through a combination of high back-drivability and rigid-flexible hybrid mechanism design (see Sec.~\ref{s:design}).
Therefore, we evaluate the robustness of our gripper by measuring its compliance characteristics: its ability to flexibly and smoothly absorb mechanical force when compressed. 
In our evaluation, we tested four different grippers: \hand{}, Franka Panda with original fingers, and Franka Panda with Fin-Ray fingers printed with 87A TPU and 95A TPU materials. The hands are placed in a device with a linear actuator that repeatedly presses them at the fingertips, measuring the resulting reactive force from the hands as a function of the applied displacement (Fig.~\ref{fig:compliance}). 
Each test was conducted six times, pushing up to \SI{40}{\milli\meter} into the resting surface of the fingertip or until a maximum reactive force of \SI{60}{\newton} was reached. 
Due to the high back-drivability of our hand, the fingers absorb some of the impact by changing its configuration: for a fair comparison, we command the gripper to return to the initial position after each collision.

Fig.~\ref{fig:compliance} (right), depicts the result of our experiment. The original Franka hand, popularly used by the robot learning community, exhibited the lowest compliance because of the rigid nature of all the components of the mechanism. 
The Franka hand with Fin-Ray fingers improves compliance and therefore, robustness, with the 95A TPU exhibiting lower compliance than the 87A due to the stiffer material. 
\hand{} achieves the highest compliance, reacting with only \SI{22}{\newton} to the maximum collision at \SI{40}{\milli\meter}.
The combination of rigid-flexible linkages and high back-drivability provides superior compliance leading to better collision robustness.
Finally, the robustness-durability of \hand{} was tested by applying repeatable \SI{40}{\milli\meter} pushing collisions to the fingertips 200 times before using it to grasp. The gripper did not show any signs of wear or tear, and operated without changes.
This demonstrates that \hand{} is highly robust and can be used in robot learning procedures involving hundreds of repetitive contact interactions (see Sec.~\ref{ss:RL}).


\begin{figure}
    \centering
\begin{subfigure}[T]{.12\textwidth}
\includegraphics[trim={0 0 0 0},clip,width=1\columnwidth]{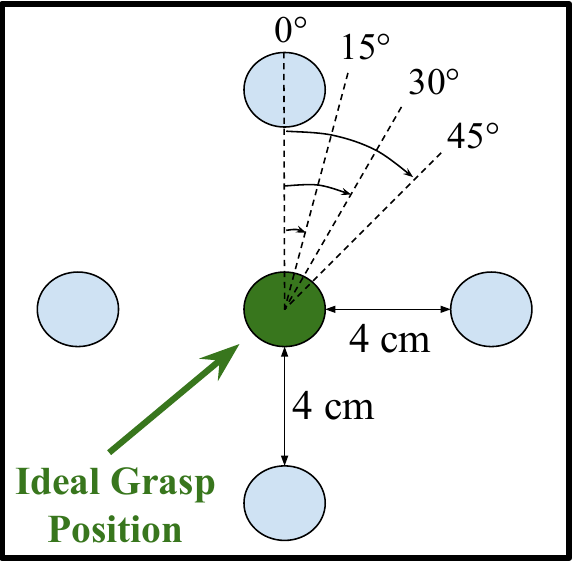}
\includegraphics[trim={0 0 0 0},clip,width=1\columnwidth]{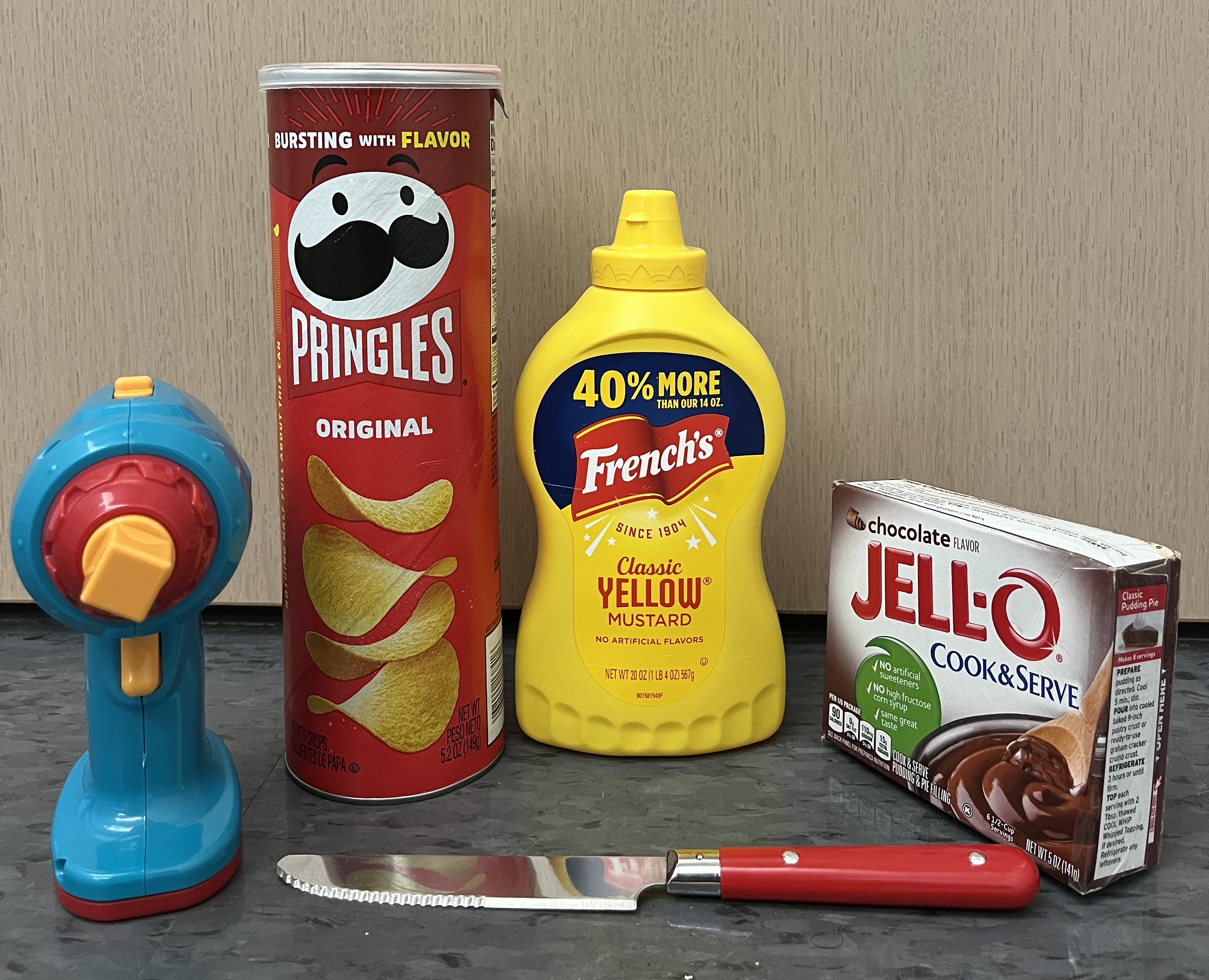}
\end{subfigure}
\begin{subfigure}[T]{.35\textwidth}
    \includegraphics[trim={0 0 0 2cm},clip,width=1.1\columnwidth]{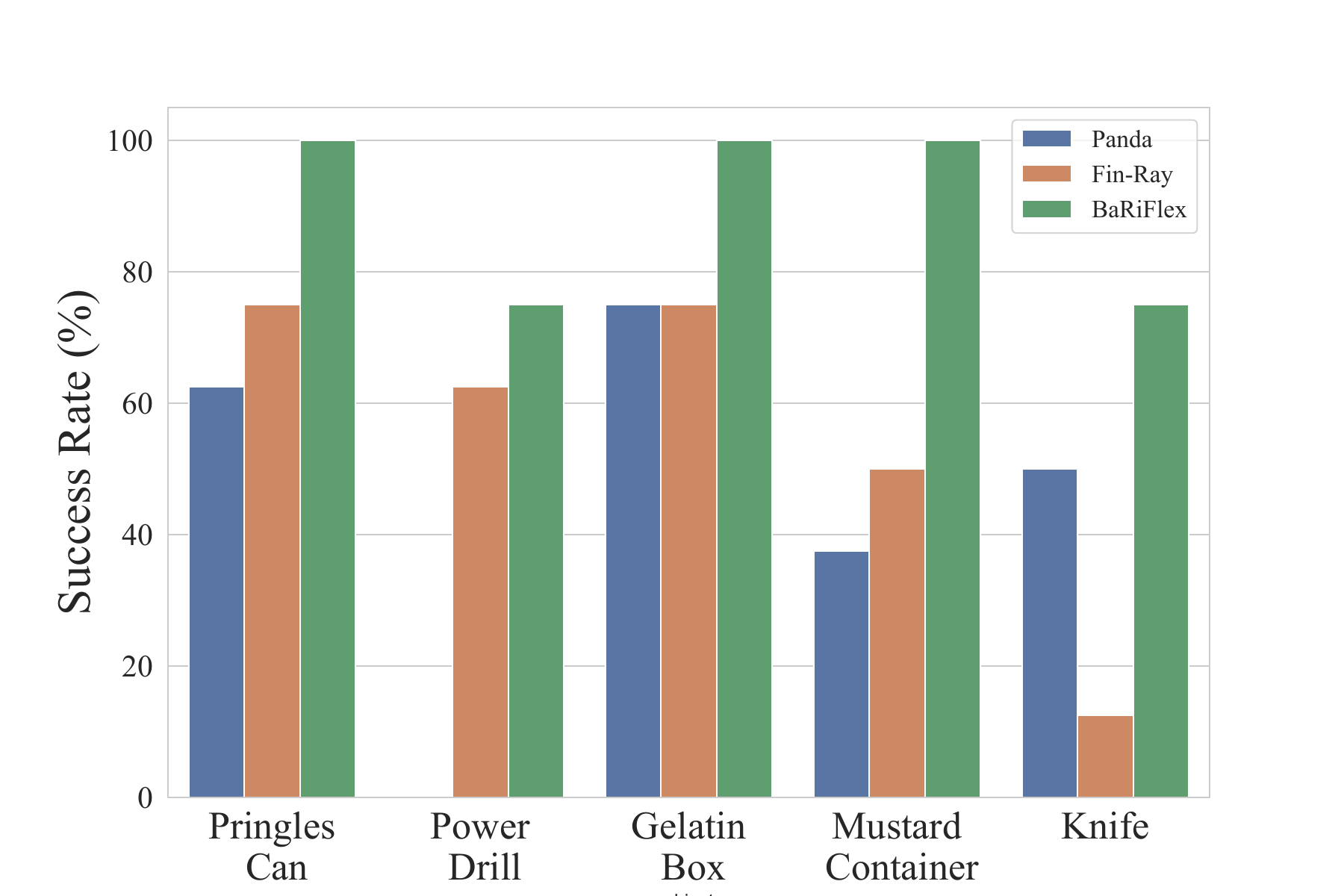}
    \label{fig:gv_results}
\end{subfigure}
    \caption{Experimental evaluation for grasp versatility. (Top-left) target objects are placed at five locations and with four orientations to evaluate the tolerance of the grippers to inaccuracies in poses. (Bottom-left) We test multiple objects with different shapes, sizes, surfaces, and weights. (Right) Results: \hand{} demonstrates a superior grasp versatility in all cases, especially with objects that require pinch grasps or conformity, thanks to its rigid-flexible design.}
    \label{fig:gv}
\end{figure}

\subsection{Evaluating Grasping Versatility}
To test \hand{}'s ability to grasp objects of various shapes and sizes, we used a subset of YCB objects~\cite{calli_ycb_2015} -- pringles can, mustard bottle, power drill, knife and gelatin box -- as shown in Fig. \ref{fig:gv}. We assessed the tolerance of \hand{} and other hands (Franka, and Franka with Fin-Ray fingers) to inaccurate expected poses of the objects. To that end, we designed an experiment where we considered five distinct positions, each located \SI{4}{\centi\meter} away from the expected position (see Fig.~\ref{fig:gv} top-left). We also tested four different orientations at 0, 15, 30, and 45 degrees from the canonical orientation, defined as the orientation that offers the thinner object side to the gripper aperture. The gripper is mounted on a Panda robot arm that moves it along a pre-defined trajectory until a constant pose where the gripper closes and then the hand moves up. If the object is lifted, the grasp is successful. 

Fig.~\ref{fig:gv} (right) summarizes the results of our experiments. We observe that \hand{} outperforms the other grippers for all objects. 
In particular, comparing the power drill and knife reveals intriguing insights. The panda gripper performs well with the knife, owing to its pinch grasp capability, while the Fin-Ray gripper handles the power drill effectively due to its conforming ability. In contrast, \hand{} stands out as the top performer, combining conformity to various shapes through its rigid-flexible construction and precise pinch grasp.
Analyzing successes per location and orientation, we also observe a superior performance of \hand{}. 
In particular, the Panda Gripper faces difficulties with orientation deviations due to its small fingers and lack of flexible material, while the Fin-Ray gripper encounters issues with distant positions because its fingertips can't exert sufficient force to maintain the grasp.
Conversely, The morphology of \hand{} with larger fingers and caging area increases its tolerance to variations in position and orientation.

\begin{figure}[t]
\begin{subfigure}[t]{1.0\columnwidth}
\includegraphics[trim={5cm 0 5cm 1cm},clip,width=0.19\textwidth]{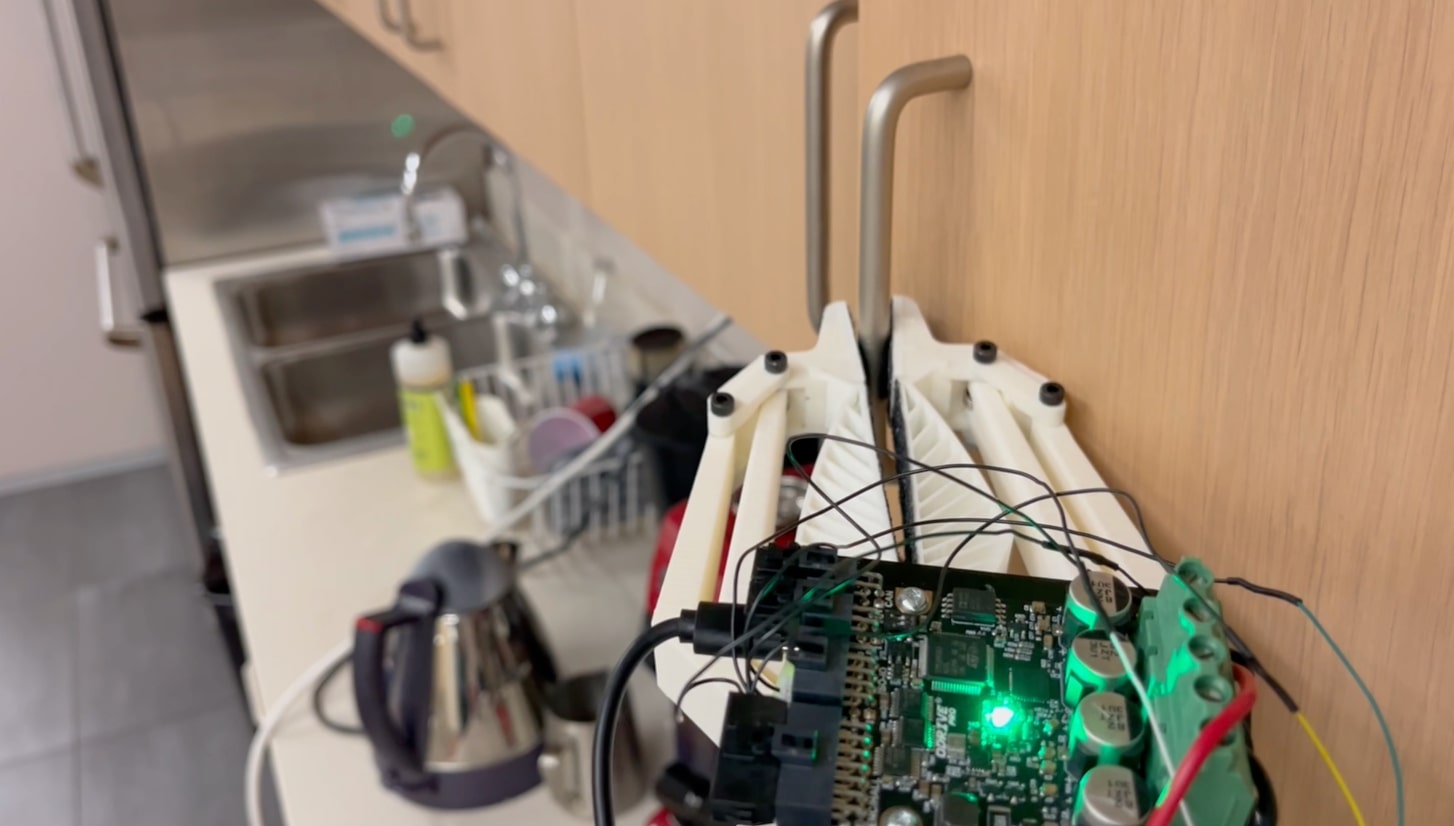}
\includegraphics[trim={2cm 0 7cm 0},clip,width=0.19\textwidth]{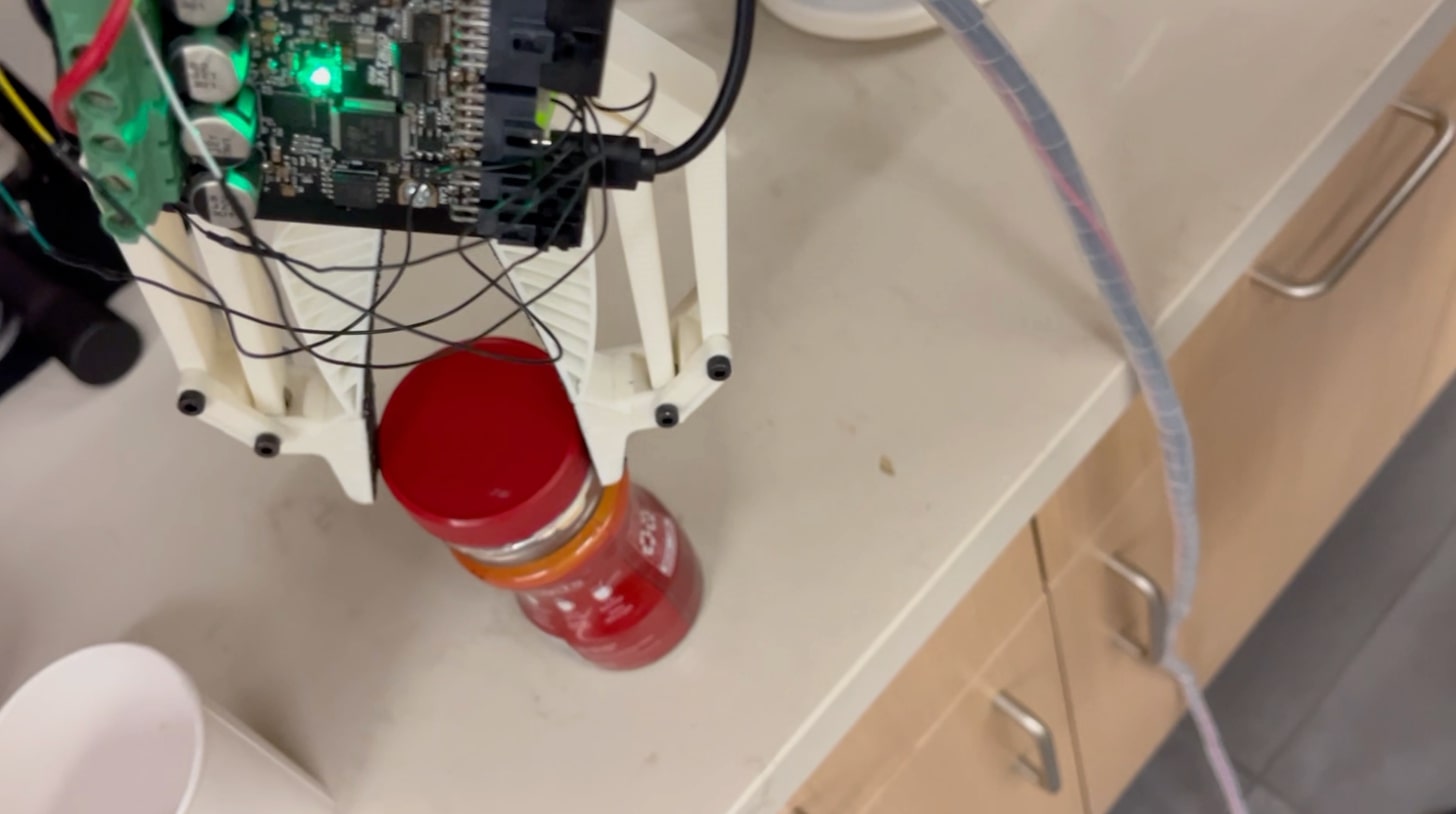}
\includegraphics[trim={5cm 3cm 5cm 0},clip,width=0.19\textwidth]{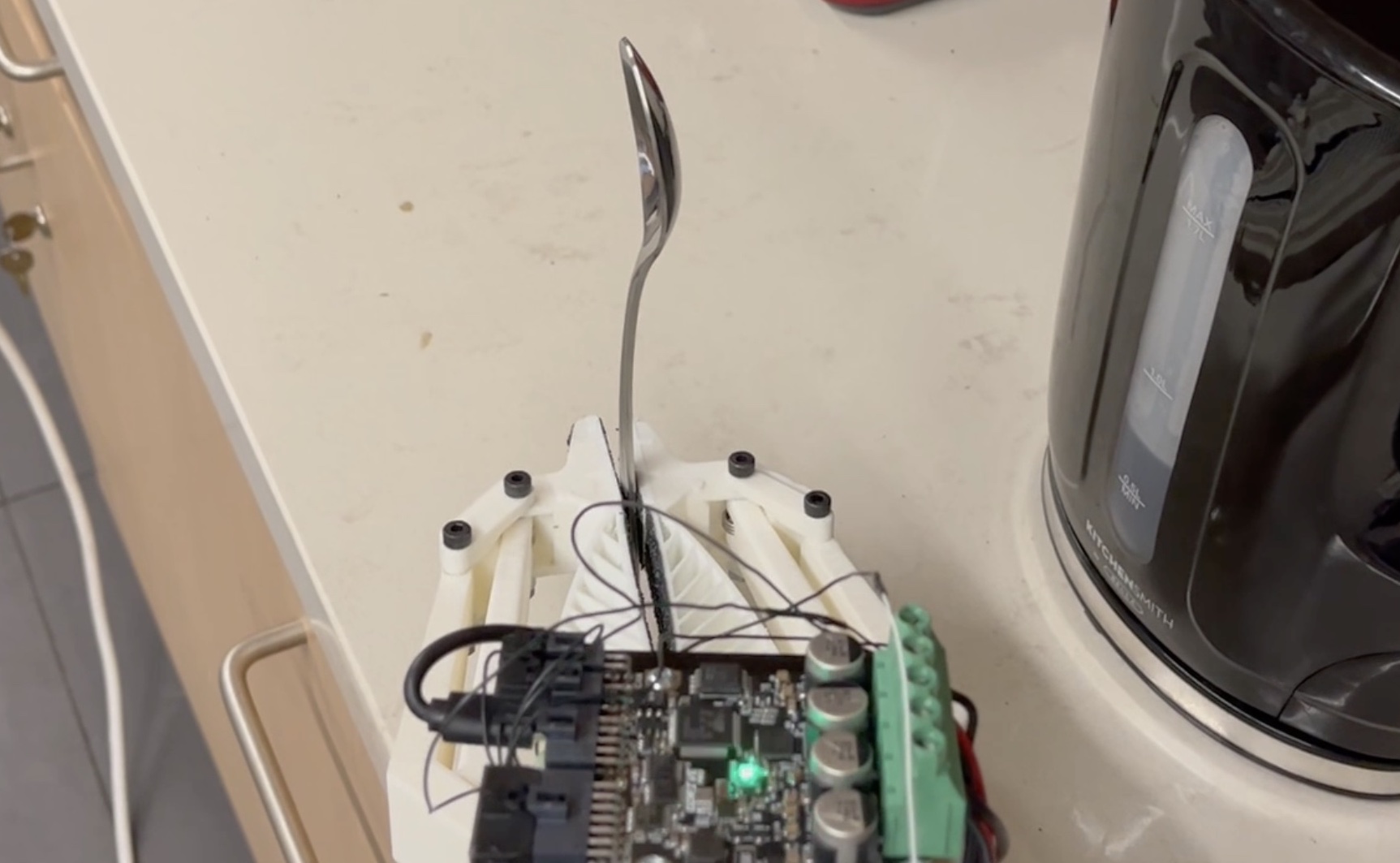}
\includegraphics[trim={7cm 0 2cm 0},clip,width=0.19\textwidth]{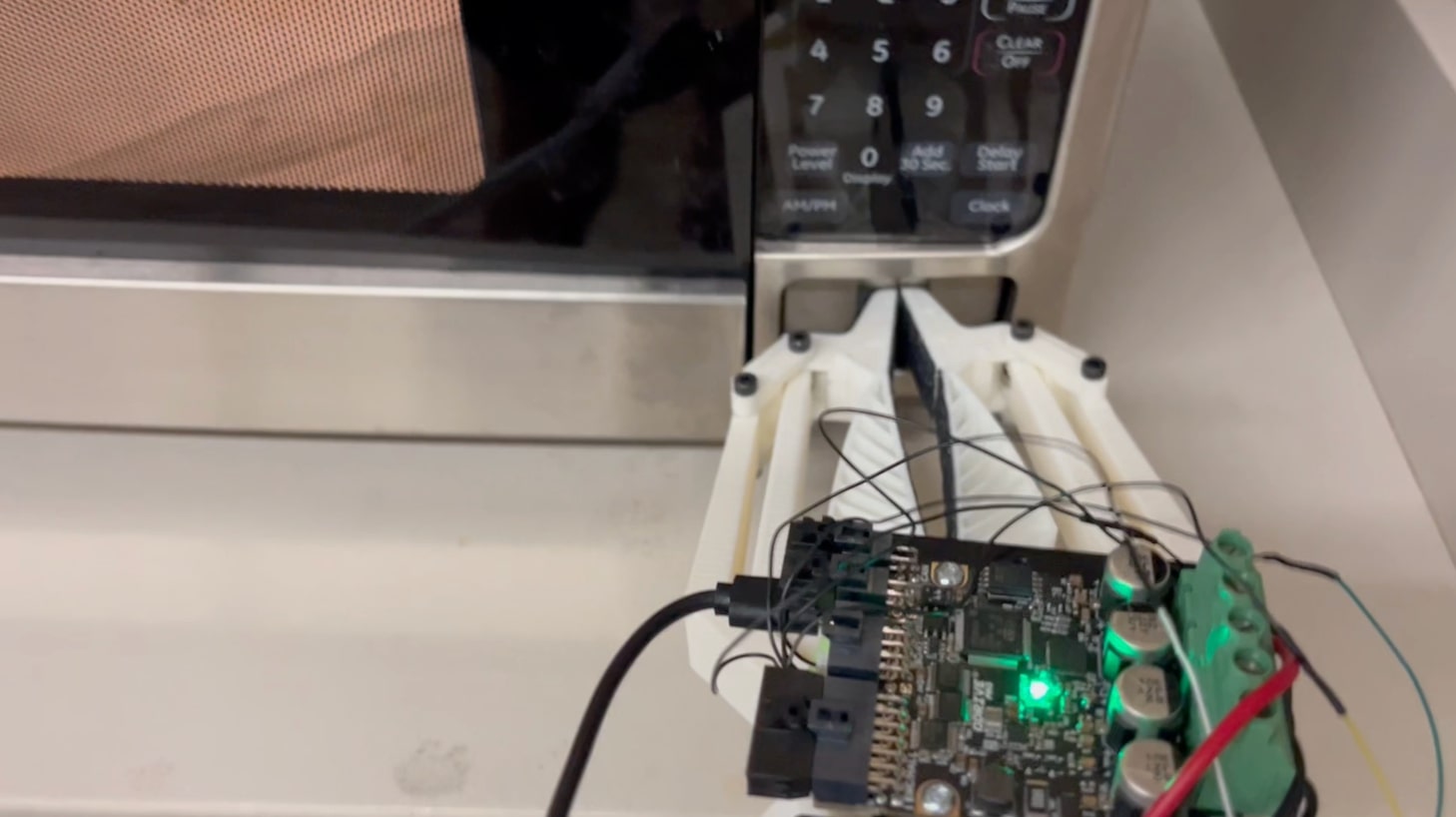}
\includegraphics[trim={5cm 0 5cm 0},clip,width=0.19\textwidth]{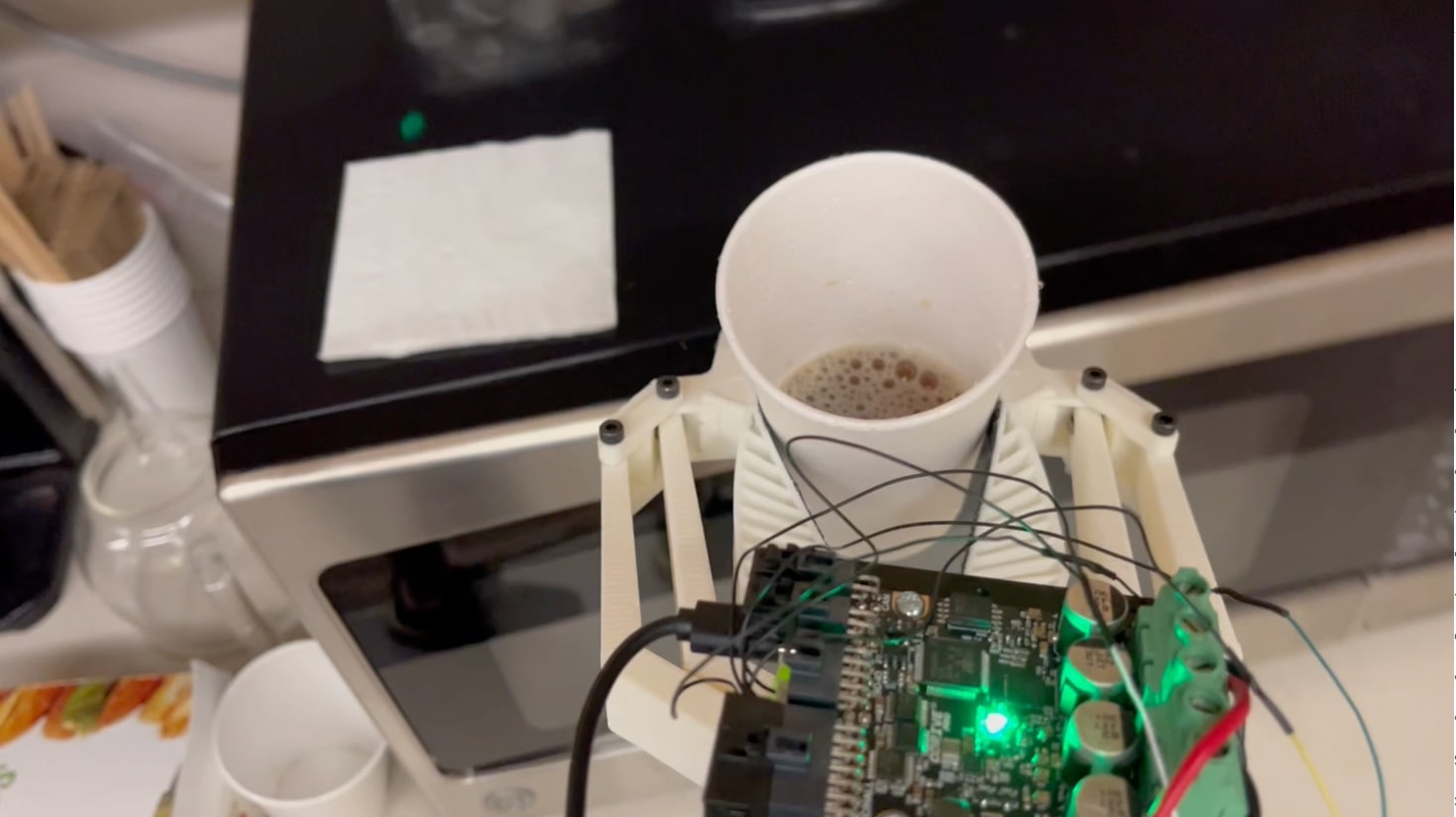}
\end{subfigure}
\\[-2ex]
\begin{subfigure}[T]{0.84\columnwidth}
\vspace{-2pt}
\includegraphics[trim={0 0 0 0},clip,width=.475\textwidth]{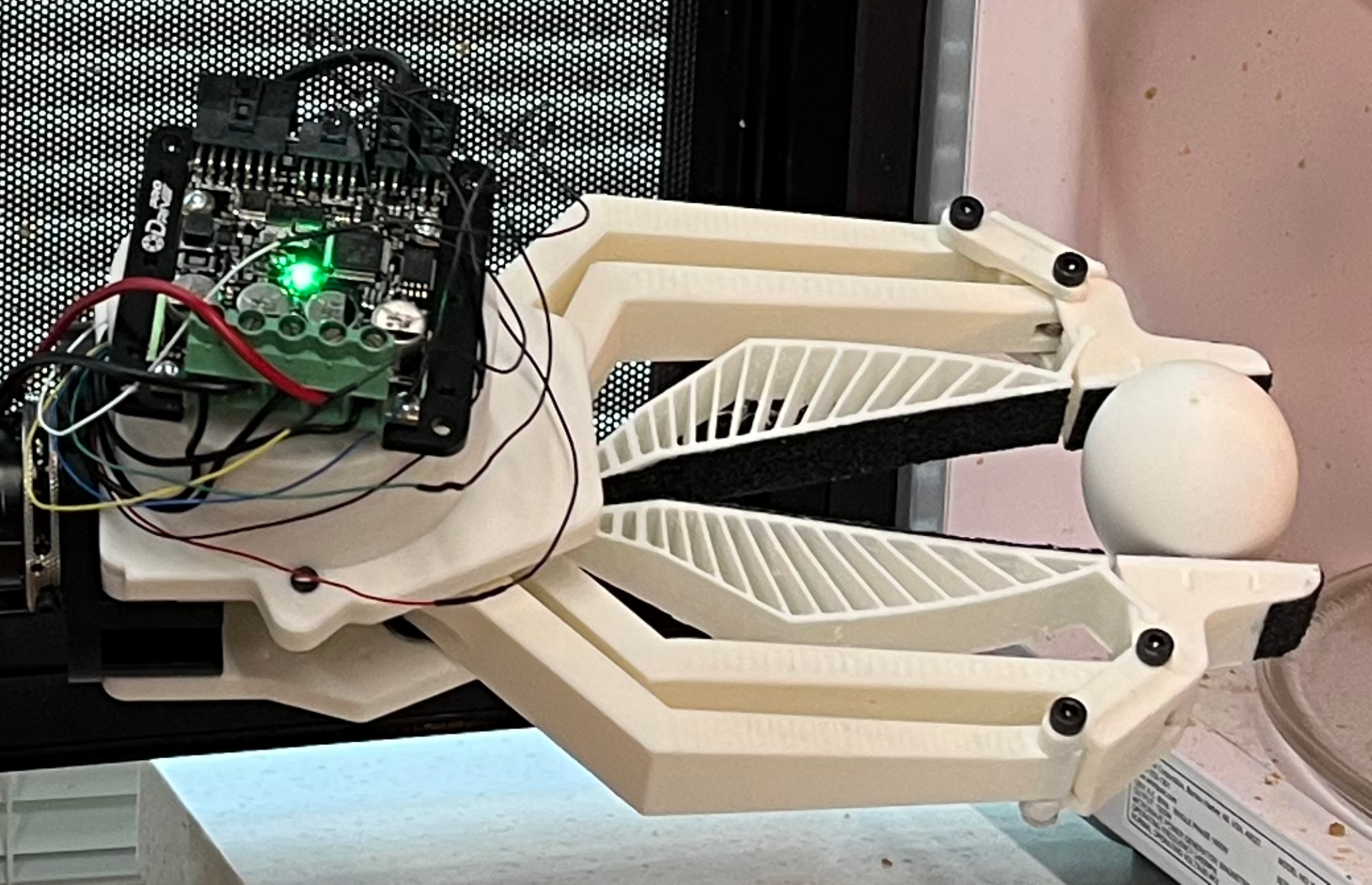}
\hfill
\includegraphics[trim={0 0 0 0},clip,width=.505\textwidth]{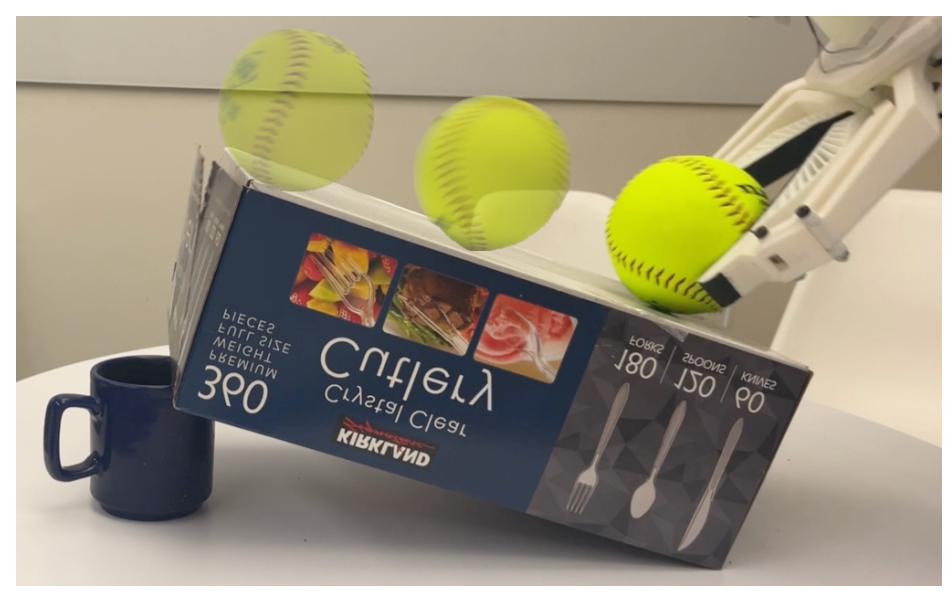}
\end{subfigure}
\begin{subfigure}[T]{0.143\columnwidth}
\includegraphics[trim={0 0 0 0},clip,angle=90,origin=c,width=1\textwidth]{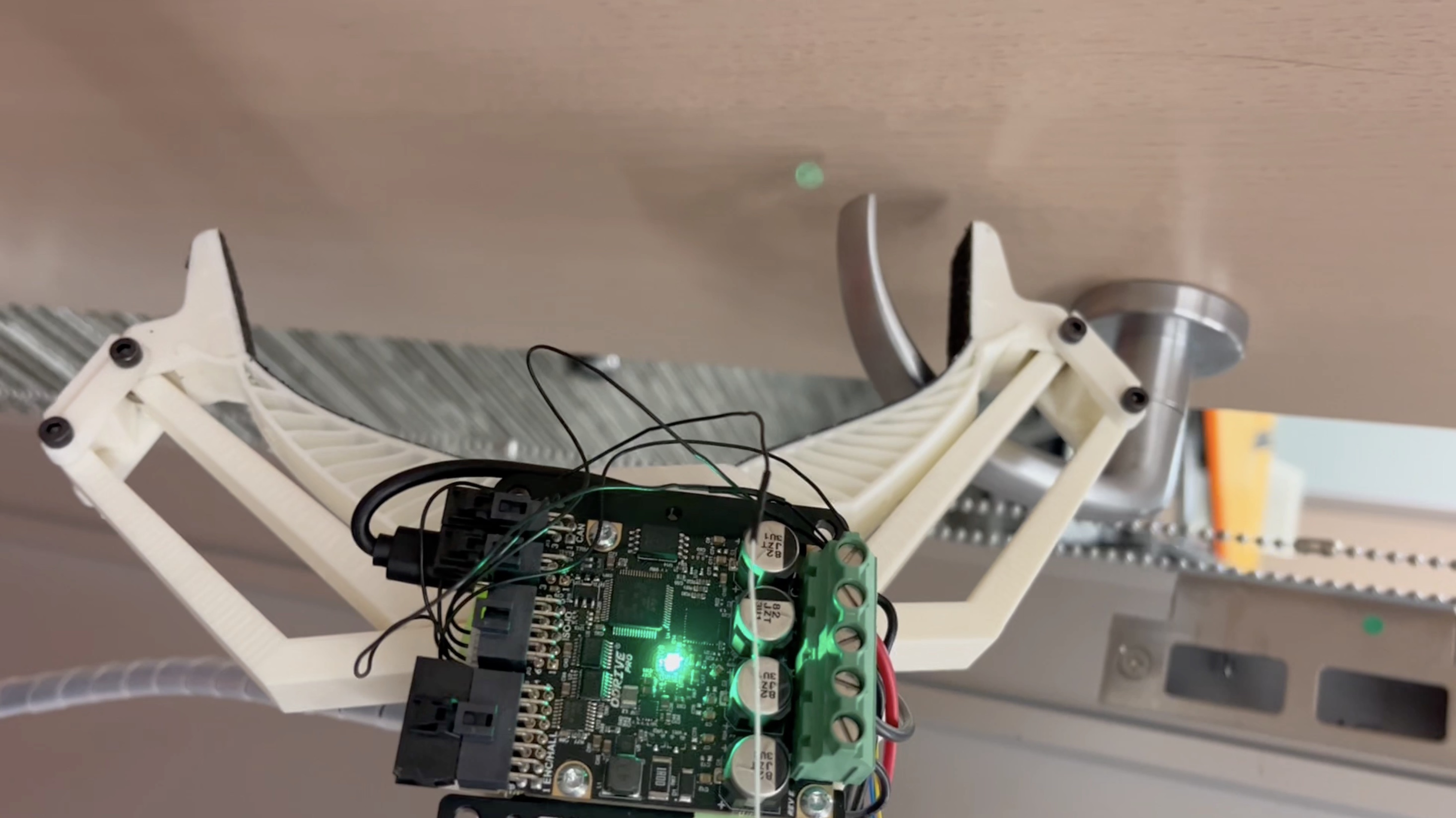}
\end{subfigure}
    \caption{Experimental evaluation for task versatility. \hand{} is mounted on a portable device and used by a human to perform multiple household tasks including a full long horizon activity, \textit{prepare a cup of instant coffee} that involves tasks such as (top row, left-to-right): opening cabinets, opening jars, pinch-grasping utensils, pressing buttons, grasping delicate plastic cups. We also evaluated other tasks such as grasping an egg, catching a fast-moving object, and opening heavy doors (Bottom row). Through this human-controlled manipulation, we empirically evidence the high task-variability enabled by \hand{}, which covers a significant fraction of possible tasks in household domains.}
    \label{fig:tv}

\end{figure}

\subsection{Evaluating Task Versatility}

We perform a qualitative evaluation of \hand{}'s ability to enable performing a wide range of tasks in household domains. 
For this evaluation, \hand{} is mounted on a handheld portable device and is operated by a human.
Fig.~\ref{fig:tv} summarizes some of the tasks tested that \hand{} is able to open cabinets and heavy doors, grasp thin objects such as spoons, open the cap of a coffee bottle, and press buttons on a microwave.

We also perform a dedicated test to evaluate the grasp precision of BaRiFlex particularly for delicate objects. The test involves the gripper's fingertip pressing a dial indicator with a resolution of 0.001 inches, 25 times. We measure the pressing displacement.
The results indicate that the average pressing displacement is \SI{3.7597}{\milli\meter} with a standard deviation of \SI{0.0253}{\milli\meter} and a maximum deviation of just \SI{0.0889}{\milli\meter}. This exceptional precision is attributed to the gripper's utilization of a rigid, low-friction four-bar linkage mechanism and a high-precision Direct-Drive actuator, which collectively enable the precise and consistent motion required for accurate object manipulation and grasping of delicate objects. The utility of \hand{}'s high precision is evaluated qualitatively through tasks such as grasping an egg (Fig.~\ref{fig:tv} bottom-left). 
Lastly, the direct-drive actuator of \hand{} enables high-speed actuation that can be critical for reactive tasks in unstructured environments (e.g., grasping a falling object). We evaluate this ability with an experiment where a ball slides along a slope and is grasped by the gripper (Fig.~\ref{fig:tv} bottom-centre).
The large variability of tasks supported by \hand{} is a significant departure from specialized grippers; while only controlled manually, our experiments evidence that \hand{} supports general manipulation in unstructured environments for household application domains.


\subsection{\hand{} for Robot Learning}
\label{ss:RL}
\begin{figure}
    \centering
    \begin{subfigure}[T]{.12\textwidth}
\includegraphics[trim={0 0 0 0},clip,width=1\columnwidth]{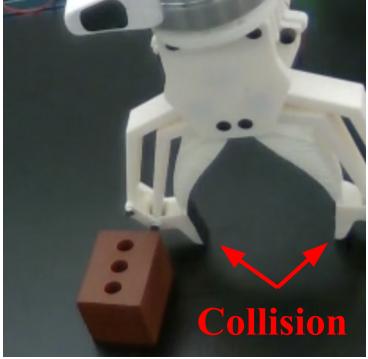}
\includegraphics[trim={0 0 0 0},clip,width=1\columnwidth]{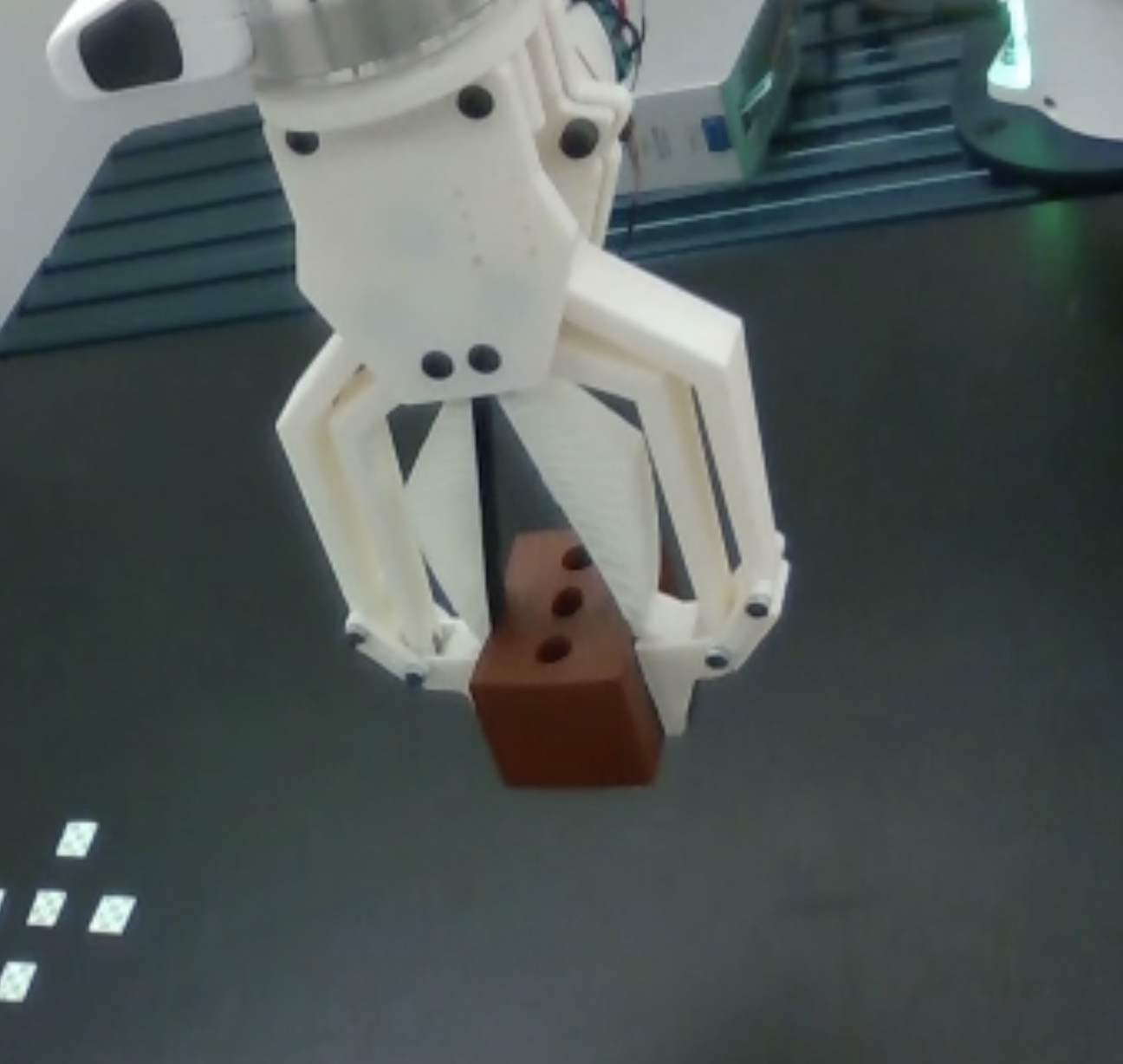}
\end{subfigure}
\begin{subfigure}[T]{.35\textwidth}
\includegraphics[trim={0 0 0 2cm},clip,width=1\columnwidth]{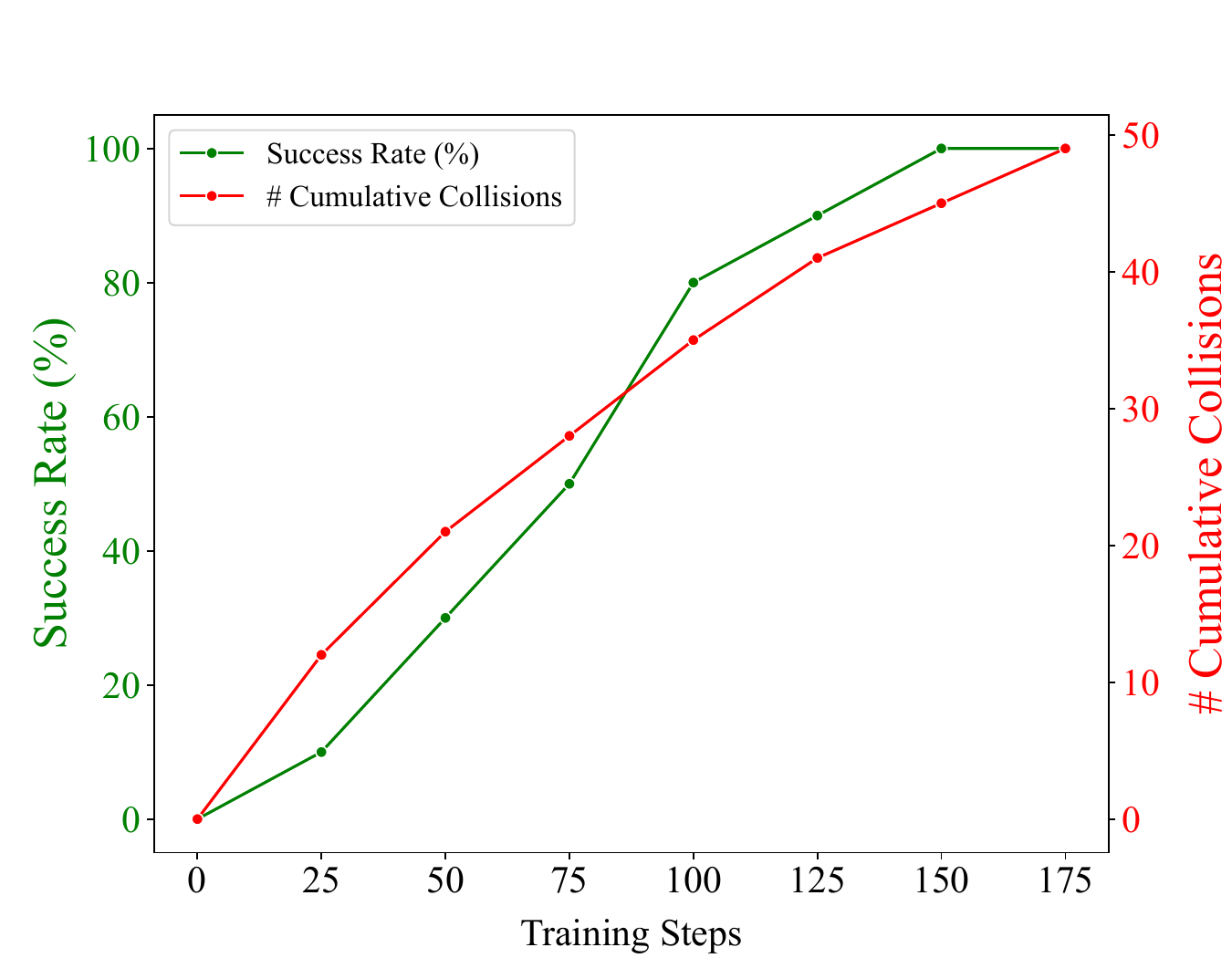}
\end{subfigure}
    \caption{Experimental evaluation for \hand{} supporting real-world reinforcement learning. The hand is used to learn to pick up a cube from trial and error. The gripper collides with the surface a total of 49 times without any damage eventually achieving a 100\% success rate at the task. (top-left) shows an example of collision and (bottom-left) shows an example of a successful grasp. The robustness and versatility of \hand{} enable the multiple unexpected collisions involved in learning manipulation tasks in unstructured environments.}
    \label{fig:Learning_Test}
\end{figure}

Our ultimate goal in developing \hand{} is to enable safe and efficient robot learning in the real world. We designed a final experiment where we evaluated the capabilities of \hand{} to operate under repeated collisions during the training process of a task using reinforcement learning. The task is to learn to grasp a cube placed on a table. 
We use Upper Confidence Bound~\cite{auer_2002} as the algorithm to learn this task with each epoch consisting of 15 train steps and 8 evaluation steps. 
The algorithm consists of actions with varying levels of contact with the tabletop surface. 
Fig. \ref{fig:Learning_Test} depicts the results of the training process and the number of collisions withstood by our hand. \hand{} absorbs up to 49 collisions without any damage or abrupt halting of the robot, enabling learning. The final policy after 175 steps reaches a 100\% success rate. Thanks to the novel design in \hand{} that combines a highly back-drivable actuator and a hybrid rigid-flexible construction, real-world reinforcement learning is safe and successful for our robot without any interruptions due to the numerous collisions.

\section{Conclusions and Limitations}
We have presented a new approach for designing robotic grippers tailored for robot learning with an ability to manipulate in daily unstructured environments and we designed a novel gripper \hand{} based on these principles. Using a highly back-drivable actuator and a hybrid rigid-flexible mechanism we are able to achieve high robustness, grasping and task versatility while maintaining simplicity. This is evidenced by our comprehensive experiments which also include a real-world reinforcement learning test.
Some limitations of our gripper include its low in-hand dexterity due to having only two fingers since we primed simplicity over dexterity. \hand{} may also face difficulties when lifting or pulling from heavy objects with smooth surfaces that can slip out of hand. Integrating a stronger actuator or adding a rubber/silicone coat to the fingers would alleviate this problem.




\printbibliography



\end{document}